\documentclass{article}

\PassOptionsToPackage{numbers, compress}{natbib}
\usepackage[final]{neurips_2023} 
\usepackage{fancyhdr}
\usepackage[utf8]{inputenc} 
\usepackage[T1]{fontenc}    
\usepackage{microtype}      
\usepackage{capt-of}
\usepackage{xcolor}
\usepackage{xspace}
\usepackage{amsmath,amssymb,amsfonts,dsfont,pifont,bm,bbm,mathrsfs,mathtools,nicefrac}
\usepackage{algorithm,algpseudocode,listings}
\usepackage{booktabs,multirow,adjustbox,diagbox,threeparttable}
\definecolor{citeblue}{RGB}{48,111,186}
\usepackage[pagebackref=false,breaklinks=true,colorlinks=true,citecolor=citeblue,bookmarks=false]{hyperref}
\usepackage{cleveref}  

\usepackage{graphicx}
\usepackage{makecell}
\usepackage{enumitem}
\usepackage{color, colortbl}
\usepackage{tikz}
\usepackage{subcaption}
\usepackage{wrapfig}

\renewcommand{\headrule}{%
    \vbox to 0pt{%
        \hbox to\headwidth{%
            \hspace*{0.0\headwidth}%
            \rule{0.925\headwidth}{\headrulewidth}%
        }%
        \vss
    }%
}

\crefname{section}{Sec.}{Secs.}
\Crefname{section}{Section}{Sections}
\crefname{table}{Tab.}{Tabs.}
\Crefname{table}{Table}{Tables}
\crefname{figure}{Fig.}{Figs.}
\Crefname{figure}{Figure}{Figures}
\crefname{equation}{Eq.}{Eqs.}
\Crefname{equation}{Equation}{Equations}
\Crefname{appendix}{Appendix}{Appendices}
\crefname{algorithm}{Alg.}{Algs.}
\Crefname{algorithm}{Algorithm}{Algorithms}
\newcommand{\op}{{\color[RGB]{0,0,0}\text{OP}}\xspace}

\newcommand{\restuner}{{\color[RGB]{0,0,0}\texttt{Res-Tuner}}\xspace}

\definecolor{tabhighlight}{HTML}{e5e5e5}
\definecolor{natural}{HTML}{648FFF}
\definecolor{specialized}{HTML}{DC267F}
\definecolor{structured}{HTML}{362682}
\definecolor{vtabmean}{HTML}{FE6100}
\definecolor{vtabparam}{HTML}{FE6100}
\definecolor{baselinecolor}{gray}{.9}

\newcommand{\resattn}{Res-Attn }

\title{\resattn: An Enhanced Res-Tuning Approach with Lightweight Attention Mechanism}

\author{%
   Chaojie Mao \quad Zeyinzi Jiang \quad \\
   Alibaba Group \quad \\
  \texttt{\{chaojie.mcj, zeyinzi.jzyz\}@alibaba-inc.com}
}

\begin{document}
\maketitle
\thispagestyle{fancy}
\fancyhead[L]{Technical Report}
\fancyfoot[C]{\thepage}
\renewcommand{\headrulewidth}{0.4pt} %
\begin{abstract}

Res-Tuning introduces a flexible and efficient paradigm for model tuning, showing that tuners decoupled from the backbone network can achieve performance comparable to traditional methods. Existing methods commonly construct the tuner as a set of trainable low-rank decomposition matrices, positing that a low-rank subspace suffices for adapting pre-trained foundational models to new scenarios. In this work, we present an advanced, efficient tuner augmented with low-rank attention, termed \resattn, which also adheres to the Res-Tuning framework. \resattn utilizes a parallel multi-head attention module equipped with low-rank projections for query, key, and value to execute streamlined attention operations. Through training this lightweight attention module, \resattn facilitates adaptation to new scenarios. Our extensive experiments across a range of discriminative and generative tasks showcase the superior performance of our method when compared to existing alternatives.
\end{abstract}
\section{Introduction} \label{sec:intro}

Efficient tuning of pre-trained foundational models has garnered considerable interest in the research community in recent years. Res-Tuning ~\cite{restuning} introduces a unified paradigm for efficient tuning techniques, enabling flexible incorporation of various modules as the Res-Tuner beyond just low-rank matrices and adapters. It has been applied to both discriminative and generative tasks, showcasing its superiority over existing alternatives in terms of both efficacy and efficiency.

In Res-Tuning, the equivalent unbound forms of prefix tuning and prompt tuning are conceptualized as a parallel lightweight attention module, featuring trainable keys and values, and prompt embeddings, respectively. However, Res-Tuning does not delve into the variations of this lightweight attention module, particularly the effects of incorporating low-rank projections for keys and values. To complement Res-Tuning, we introduce a novel and effective tuning method called \resattn, which is based on a further analysis of the aforementioned form variation. Specifically, \resattn is designed with an unbound lightweight attention module that uses low-rank projections to get query, key, and value with the original features derived from the backbone model. Unlike parallel prompt tuning, which employs a shared linear projection with the backbone, the low-rank projection in our approach consists of a trainable linear operation. In the remainder of this technical report, we start with a comprehensive description of the proposed \resattn and proceed with experiments carried out on a variety of discriminative and generative tasks, adhering to the experimental framework established in Res-Tuning.

\section{Method} \label{sec:restuning}

\begin{figure*}[htb]
    \includegraphics[width=1.0\linewidth]{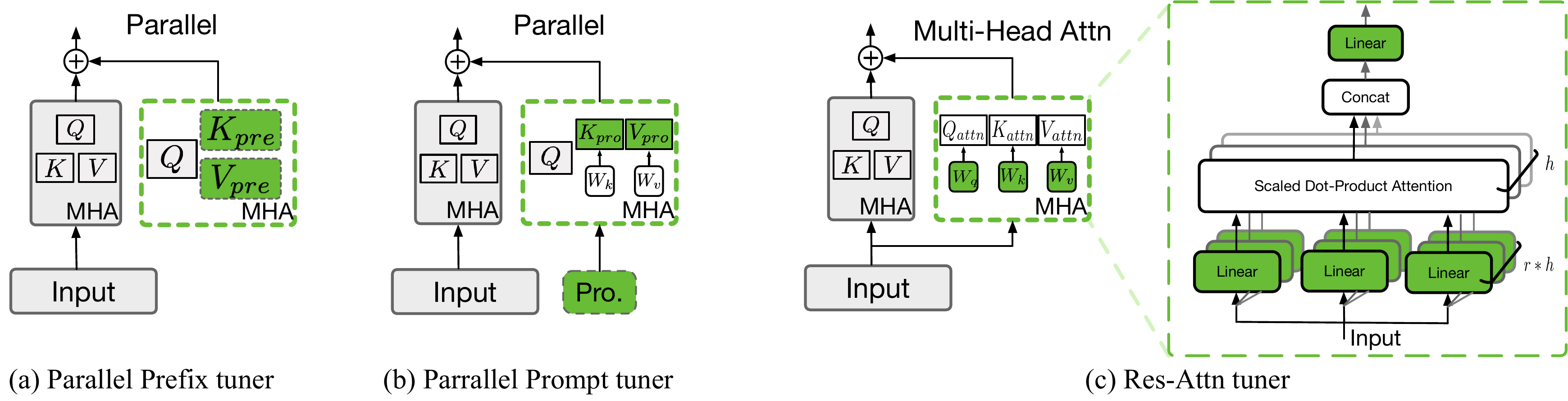}
    \caption{The structure of (a) Parallel Prefix tuner, (b) Parallel Prompt tuner, and (c) \resattn tuner. The right of (c) is the detailed structure of \resattn tuner, in which the input is projected to the output query, key, and value with the shape of $\text{rank}*\text{head}$ using the trainable linear operation.}
    \label{fig:res_tuning_attn}
\end{figure*}

\begin{algorithm*}[ht]
\caption{Implementation of \resattn tuner in PyTorch-like style.}\label{alg:code}
\definecolor{codeblue}{rgb}{0.25,0.5,0.5}
\definecolor{colorred}{RGB}{197, 49, 124}
\lstset{
  backgroundcolor=\color{white},
  basicstyle=\fontsize{8.8pt}{8.8pt}\ttfamily\selectfont,
  columns=fullflexible,
  breaklines=true,
  captionpos=b,
  commentstyle=\fontsize{7.2pt}{7.2pt}\color{codeblue},
  keywordstyle=\fontsize{7.2pt}{7.2pt}\color{colorred},
}
\begin{lstlisting}[language=python]
class ResAttn(nn.Module):
    def __init__(self, dim, rank=4, num_heads=4, qkv_bias=False, attn_drop=0., proj_drop=0.):
        super().__init__()
        self.num_heads = num_heads
        self.rank = rank
        self.scale = rank ** -0.5
        self.qkv = nn.Linear(dim, self.rank * num_heads * 3, bias=qkv_bias)
        self.attn_drop = nn.Dropout(attn_drop)
        self.o = nn.Linear(self.rank * num_heads, dim)
        self.o_drop = nn.Dropout(proj_drop)
        self._kaiming_init_weights(self.qkv)
        self._zero_init_weights(self.o)
    def _zero_init_weights(self, m):
        if isinstance(m, nn.Linear):
            nn.init.zeros_(m.weight)
            nn.init.zeros_(m.bias)
    def _kaiming_init_weights(self, m):
        if isinstance(m, nn.Linear):
            nn.init.kaiming_uniform_(m.weight, a=math.sqrt(5))
    def forward(self, x):
        B, N, C = x.shape
        qkv = self.qkv(x).reshape(B, N, 3, self.num_heads, self.rank).permute(2, 0, 3, 1, 4)
        q, k, v = qkv.unbind(0)
        attn = (q @ k.transpose(-2, -1)) * self.scale
        attn = attn.softmax(dim=-1)
        attn = self.attn_drop(attn)
        x = (attn @ v).transpose(1, 2).reshape(B, N, self.num_heads*self.rank)
        x = self.o(x)
        x = self.o_drop(x)
        return x
\end{lstlisting}
\end{algorithm*}

In this section, we provide a detailed description of the \resattn tuner. Res-Tuning ~\cite{restuning} is derived from a unified formula that combines a frozen pre-trained operation with a tuner composed of learnable parameters, as follows:
\begin{equation}
\label{eq:uni}
\boldsymbol{x}^{\prime} = \op(\boldsymbol{x})  + \restuner(\boldsymbol{x}) ,
\end{equation}
where $\op$ denotes existing operations in the pre-trained backbone such as Multi-Head Attention (MHA) and Feed-Forward Network (FFN), while the \restuner\ represents the learnable structures that are connected in parallel to the existing operations. 

The Parallel Prefix tuner and Parallel Prompt tuner are the tuning mechanisms introduced in Res-Tuning as part of the proposed \restuner framework. As illustrated in ~\cref{fig:res_tuning_attn}{\color{red}a}, the Parallel Prefix tuner executes an independent attention process utilizing trainable parameters as key and value, while reutilizing the query from the attention module of the main backbone. As shown in ~\cref{fig:res_tuning_attn}{\color{red}b}, the Parallel Prompt tuner operates an independent attention mechanism that reuses the query from the main backbone. It employs a trainable prompt vector, which is projected into key and value using the shared MHA linear projection from the main backbone.

Drawing inspiration from the aforementioned tuners, the \resattn tuner utilizes a fully independent lightweight self-attention mechanism. It features trainable MHA parameters while sharing the input from the main backbone's MHA. As depicted in ~\cref{fig:res_tuning_attn}{\color{red}c}, the input from the main backbone undergoes a trainable linear projection to produce the low-rank query, key, and value matrices $\in \mathbb{R}^{r\times h}$, where $r$ represents the rank of the low-rank projection, and $h$ denotes the number of heads. The forward function implementation of \resattn tuner is provided in ~\cref{alg:code}.

\section{Experiments} 

Our evaluation of the \resattn tuner spans on transfer learning in both discriminative and generative tasks. We carry out the assessments under the same experimental conditions as those outlined in Res-Tuning ~\cite{restuning}. In our experimental work on discriminative tasks using CIFAR-100 and VTAB-1K datasets, we utilize the ViT-B/16 model pre-trained on ImageNet21K as our backbone architecture. For the generative tasks, we employ version 1.5 of the Stable Diffusion (SD) model, conducting experiments on the few-shot anime dataset.

\begin{table}[t]
\centering
\caption{
  \textbf{Exploration of various combinations} of operations in the pre-trained backbone and various \restuner{s} achieves a stronger performance than the existing tuning strategies on CIFAR-100.
  Adapter, Prefix, Prompt, and Attention are abbreviated as Ada., Pre., Pro., and Attn., respectively.
}
\begin{subtable}[t]{.23\linewidth}
\setlength{\tabcolsep}{5pt}
\caption{\texttt{Ablation of $r \times h$.}}
\label{taba:ablation}
\scalebox{0.7}{
    \begin{tabular}{l|cc}
        \toprule
        MHA$\backslash$Tuner & Param. & Acc. \\
        \midrule
        $\text{Res-Attn.}_{8\times8}$ & 2.35M & 92.20  \\
        $\text{Res-Attn.}_{8\times4}$ & 1.22M & 92.51 \\
        $\text{Res-Attn.}_{4\times4}$ & 0.66M & \textbf{92.70} \\  
        $\text{Res-Attn.}_{2\times4}$ & 0.32M & 92.58 \\  
        \bottomrule
    \end{tabular}}
\end{subtable}
\hfill
\begin{subtable}[t]{.26\linewidth}
\setlength{\tabcolsep}{5pt}
\caption{\texttt{Single-Res-Tuner.}}
\label{taba:single}
\scalebox{0.7}{
            \begin{tabular}{l|ccc}
            \toprule
            Tuner$\backslash$OP & MHA & FFN & Block \\
            \midrule
            Res-Ada. & 92.46 & 92.34 & 92.49 \\
            Res-Pre. & 91.88 & 92.33 & 92.39 \\
            Res-Pro. & 92.24 & 92.68 & 92.16 \\
            Res-Attn. & \textbf{92.70} & \textbf{92.74} & \textbf{92.52} \\
            \bottomrule
    \end{tabular}}
\end{subtable}
\hfill
\begin{subtable}[t]{0.45\linewidth}
\setlength{\tabcolsep}{5pt}
\caption{\texttt{Dual-Res-Tuner.}}
\label{taba:dual}
    \scalebox{0.7}{
        \begin{tabular}{l|cccc}
            \toprule
            MHA$\backslash$FFN & Res-Ada. & Res-Pre. & Res-Pro.  & Res-Attn. \\
            \midrule
            Res-Ada. & 93.25 & 92.95 & 92.62 & \textbf{93.31} \\
            Res-Pre. & 93.22 & 92.38 & 92.87 & 93.24 \\
            Res-Pro. & 93.03 & 92.92 & 92.91 & 93.24 \\  
            Res-Attn. & 93.28 & 93.14 & 93.10 & 93.30 \\  
            \bottomrule
    \end{tabular}}
\end{subtable}
        
\label{taba:combinations}
\end{table}

\begin{table}[t]
\caption{
\textbf{Performance and efficiency comparison} on the VTAB-1K benchmark with ViT-B/16 models pre-trained on ImageNet-21K. ``Group Mean'' denotes the average accuracy of the three subgroups. ``All Mean'' denotes the average accuracy of 19 downstream tasks. 
}
\centering\scriptsize
\setlength{\tabcolsep}{1pt}
\begin{tabular}{l | ccccccc | cccc | cccccccc | ccccccc}
\toprule
& \multicolumn{7}{c|}{\textbf{Natural}} & \multicolumn{4}{c|}{\textbf{Specialized}} & \multicolumn{8}{c|}{\textbf{Structured}} \\
  & \rotatebox{90}{\raisebox{0.5pt}{\tikz\fill[natural] (0,0) circle (.5ex);} CIFAR-100}
 & \rotatebox{90}{\raisebox{0.5pt}{\tikz\fill[natural] (0,0) circle (.5ex);} Caltech101}
 & \rotatebox{90}{\raisebox{0.5pt}{\tikz\fill[natural] (0,0) circle (.5ex);} DTD}
 & \rotatebox{90}{\raisebox{0.5pt}{\tikz\fill[natural] (0,0) circle (.5ex);} Flowers102}
 & \rotatebox{90}{\raisebox{0.5pt}{\tikz\fill[natural] (0,0) circle (.5ex);} Pets}
 & \rotatebox{90}{\raisebox{0.5pt}{\tikz\fill[natural] (0,0) circle (.5ex);} SVHN}
 & \rotatebox{90}{\raisebox{0.5pt}{\tikz\fill[natural] (0,0) circle (.5ex);} Sun397}
 & \rotatebox{90}{\raisebox{0.5pt}{\tikz\fill[specialized] (0,0) circle (.5ex);} Camelyon}
 & \rotatebox{90}{\raisebox{0.5pt}{\tikz\fill[specialized] (0,0) circle (.5ex);} EuroSAT}
 & \rotatebox{90}{\raisebox{0.5pt}{\tikz\fill[specialized] (0,0) circle (.5ex);} Resisc45}
 & \rotatebox{90}{\raisebox{0.5pt}{\tikz\fill[specialized] (0,0) circle (.5ex);} Retinopathy}
 & \rotatebox{90}{\raisebox{0.5pt}{\tikz\fill[structured] (0,0) circle (.5ex);} Clevr-Count}
 & \rotatebox{90}{\raisebox{0.5pt}{\tikz\fill[structured] (0,0) circle (.5ex);} Clevr-Dist}
 & \rotatebox{90}{\raisebox{0.5pt}{\tikz\fill[structured] (0,0) circle (.5ex);} DMLab}
 & \rotatebox{90}{\raisebox{0.5pt}{\tikz\fill[structured] (0,0) circle (.5ex);} KITTI-Dist}
 & \rotatebox{90}{\raisebox{0.5pt}{\tikz\fill[structured] (0,0) circle (.5ex);} dSpr-Loc}
 & \rotatebox{90}{\raisebox{0.5pt}{\tikz\fill[structured] (0,0) circle (.5ex);} dSpr-Ori}
 & \rotatebox{90}{\raisebox{0.5pt}{\tikz\fill[structured] (0,0) circle (.5ex);} sNORB-Azim}
 & \rotatebox{90}{\raisebox{0.5pt}{\tikz\fill[structured] (0,0) circle (.5ex);} sNORB-Elev}
 & \rotatebox{90}{\raisebox{0.5pt}{\tikz\fill[vtabmean] (0,0) circle (.5ex);} Group Mean}  
 & \rotatebox{90}{\raisebox{0.5pt}{\tikz\fill[vtabmean] (0,0) circle (.5ex);} All Mean} 
 & \rotatebox{90}{\raisebox{0.5pt}{\tikz\fill[vtabparam] (0,0) circle (.5ex);} Param. (M)} 
 & \rotatebox{90}{\raisebox{0.5pt}{\tikz\fill[vtabparam] (0,0) circle (.5ex);} Mem. (GB)} \\
\midrule
\multicolumn{24}{c}{\emph{Traditional methods}}\\
Full & 68.9 & 87.7 & 64.3 & 97.2 & 86.9 & 87.4 & 38.8 & 79.7 & 95.7 & 84.2 & 73.9 & 56.3 & 58.6 & 41.7 & 65.5 & 57.5 & 46.7 & 25.7 & 29.1 & 68.96 & 65.57 & 85.84 & 9.40 \\
Linear & 63.4 & 85.0 & 63.2 & 97.0 & 86.3 & 36.6 & 51.0 & 78.5 & 87.5 & 68.6 & 74.0 & 34.3 & 30.6 & 33.2 & 55.4 & 12.5 & 20.0 & 9.6 & 19.2 & 57.64 & 52.94 & 0.04 & 3.09 \\
\midrule
\multicolumn{24}{c}{\emph{Parameter-efficient tuning methods}}\\
Adapter~\cite{houlsby2019adapter} & 74.2 & 85.7 & 62.7 & 97.8 & 87.2 & 36.4 & 50.7 & 76.9 & 89.2 & 73.5 & 71.6 & 45.2 & 41.8 & 31.1 & 56.4 & 30.4 & 24.6 & 13.2 & 22.0 & 60.52 & 56.35 & 1.82 & 6.53 \\
LoRA~\cite{hu2021lora} & 67.1 & 91.4 & 69.4 & 98.8 & 90.4 & 85.3 & 54.0 & 84.9 & 95.3 & 84.4 & 73.6 & \textbf{82.9} & \textbf{69.2} & 49.8 & 78.5 & 75.7 & 47.1 & 31.0 & 44.0 & 74.60 & 72.30 & 0.29 & 6.88 \\
VPT-Deep~\cite{jia2022vpt} & \textbf{78.8} & 90.8 & 65.8 & 98.0 & 88.3 & 78.1 & 49.6 & 81.8 & \textbf{96.1} & 83.4 & 68.4 & 68.5 & 60.0 & 46.5 & 72.8 & 73.6 & 47.9 & \textbf{32.9} & 37.8 & 71.96 & 69.43 & 0.60 & 8.13 \\
SSF~\cite{ssf2022} & 69.0 & 92.6 & \textbf{75.1} & \textbf{99.4} & 91.8 & \textbf{90.2} & 52.9 & 87.4 & 95.9 & 87.4 & 75.5 & 75.9 & 62.3 & \textbf{53.3} & 80.6 & 77.3 & 54.9 & 29.5 & 37.9 & 75.69 & 73.10 & 0.24 & 7.47 \\
NOAH~\cite{noah2022} & 69.6 & 92.7 & 70.2 & 99.1 & 90.4 & 86.1 & 53.7 & 84.4 & 95.4 & 83.9 & 75.8 & 82.8 & 68.9 & 49.9 & \textbf{81.7} & 81.8 & 48.3 & 32.8 & 44.2 & 75.48 & 73.25 & 0.42 & 7.27 \\
Res-Tuning & 75.2 & 92.7 & 71.9 & 99.3 & \textbf{91.9} & 86.7 & \textbf{58.5} & 86.7 & 95.6 & 85.0 & 74.6 & 80.2 & 63.6 & 50.6 & 80.2 & \textbf{85.4} & \textbf{55.7} & 31.9 & 42.0 & \textbf{76.32} & \textbf{74.10} & 0.55 & 8.95 \\
\rowcolor{tabhighlight} \resattn & 71.6 & \textbf{93.0} & 72.4 & \textbf{99.4} & 90.4 & 89.2 & 54.5 & \textbf{87.9} & 95.7 & 83.7 & \textbf{76.6} & 79.5 & 62.3 & 49.7 & 81.2 & 83.0 & 53.9 & 32.6 & \textbf{48.9} & 76.28 & 73.97 & 1.25 & 7.76 \\
\bottomrule
\end{tabular}
\label{taba:vtab1k}
\vspace{-15pt}
\end{table}

\begin{figure*}[htb]
    \includegraphics[width=1.0\linewidth]{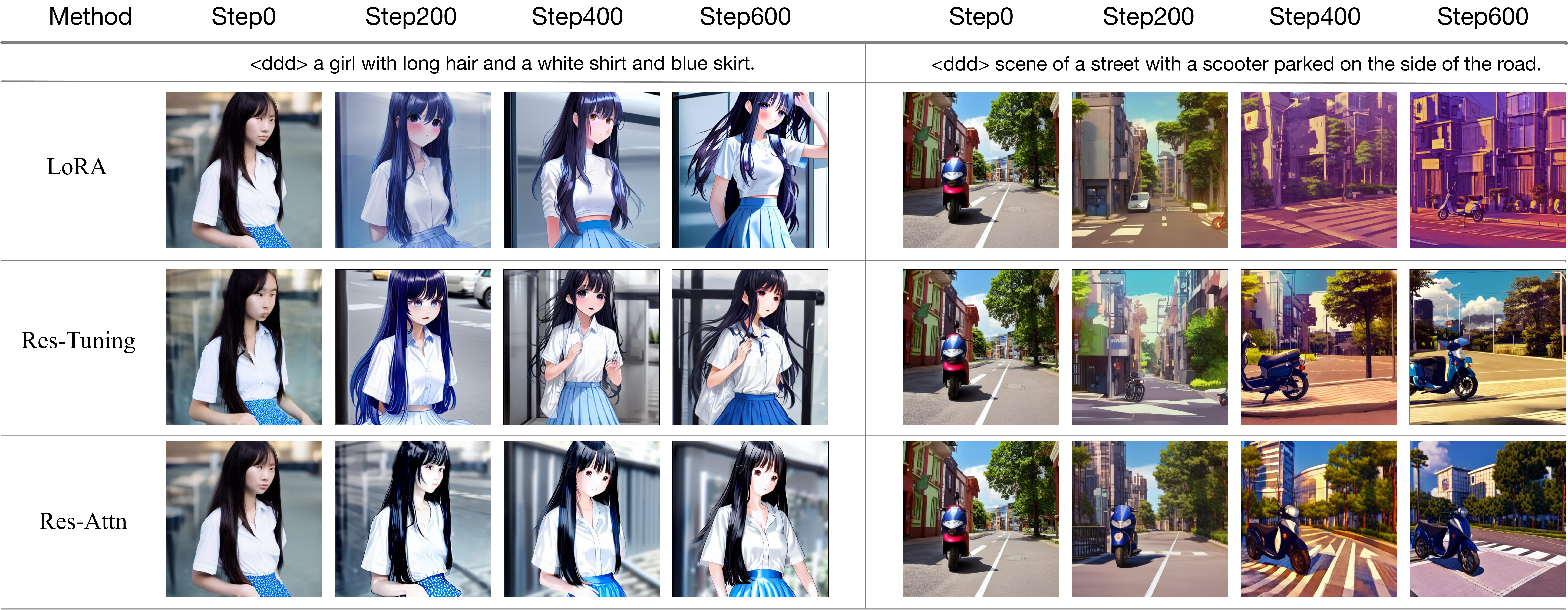}
    \caption{The comparison of performance for LoRA, Res-Tuning, and \resattn.}
    \label{fig:few_shot_anime}
\end{figure*}

\textbf{Performance of single tuning and combination of multiple tuning.} To assess the performance of the \resattn tuner relative to other tuning approaches, we begin with an ablation study focusing on the rank and the number of heads within the \resattn framework, using the CIFAR-100 dataset. As illustrated in ~\cref{taba:combinations}{\color{red}a}, the \resattn tuner reaches its optimal performance with an accuracy of 92.70\% when set to a rank $r=4$ and a head number $h=4$. Following this, we examine the impact of employing either a single tuner or a pair of tuners for each block. By adopting a rank of $r=4$ and a head number of $h=4$ as the standard configuration, we observe enhancements over other tuners. As depicted in ~\cref{taba:combinations}{\color{red}b}, the \resattn tuner attains accuracy of 92.70\%, 92.74\%, and 92.52\% when implementing a single tuner in MHA, FFN, and the entire Block, respectively. When utilizing a combination of two tuners, the \resattn tuner reaches its peak performance at 93.31\% when paired with the adapter tuner, and the pair of two \resattn tuners achieves a comparable performance as 93.30\%.

\textbf{Transfer learning on discriminative task.} Our primary evaluation focuses on the basic transfer learning scenario, in which pre-trained models are fine-tuned to adapt to various downstream tasks. VTAB-1K consists of 19 different visual classification tasks that are grouped into three categories: natural, specialized, and structured. We present the results of the 19 datasets on the VTAB-1K benchmark to evaluate the performance of the proposed \resattn tuner in ~\cref{taba:vtab1k}. \resattn secures the leading performance in 5 subdatasets and exhibits performance on par with that of Res-Tuning.


\textbf{Few-Shot learning on generative task.}  We conducted few-shot transfer learning on an anime dataset utilizing a minimal set of only 30 image-prompt pairs. In ~\cref{fig:few_shot_anime}, we present a qualitative comparison of sample outputs from Res-Tuning and our approach, using an identical training configuration. 

\section{Conclusion}\label{sec:conclusion}
In this work, we propose an improved efficient tuner following the Res-Tuning paradigm, which employs light-weight independent attention as the trainable module. The extensive experiments on both discriminative and generative tasks demonstrate the comparable performance of our method to existing alternatives.
{\small
\bibliographystyle{abbrvnat}
\bibliography{neurips_2023}
}

\end{document}